\documentclass{article}

\usepackage[utf8]{inputenc}
\usepackage[hidelinks]{hyperref}
\usepackage{url}
\usepackage{booktabs}
\usepackage{amsfonts}
\usepackage{amsmath}
\usepackage{amssymb}
\usepackage{nicefrac}
\usepackage{graphicx}
\usepackage{natbib}
\usepackage[margin=1in]{geometry}
\usepackage{enumitem}
\usepackage{xcolor}
\usepackage{colortbl}
\usepackage{multirow}
\usepackage{longtable}
\usepackage{algorithm}
\usepackage{algorithmic}
\usepackage{tcolorbox}

\definecolor{atlasgreen}{RGB}{26, 58, 26}        % Darkest army green #1A3A1A
\definecolor{atlasdark}{RGB}{45, 74, 45}         % Dark army green #2D4A2D
\definecolor{atlasmedium}{RGB}{61, 92, 61}       % Medium army green #3D5C3D
\definecolor{atlasolive}{RGB}{74, 107, 74}       % Olive green #4A6B4A
\definecolor{atlassage}{RGB}{90, 122, 90}        % Muted sage #5A7A5A
\definecolor{keybox}{RGB}{248, 248, 245}         % Off-white #F8F8F5
\definecolor{takeawayblue}{RGB}{248, 248, 245}   % Redirected to off-white for consistency

\begin{document}

% Title with horizontal rules (matching Attention Is All You Need)
\begin{center}
\rule{\textwidth}{1.5pt}
\vspace{0.3cm}

{\LARGE \bf From Perception to Action:\\Spatial AI Agents and World Models}

\vspace{0.3cm}
\rule{\textwidth}{0.5pt}
\vspace{0.8cm}

% Authors in grid layout
\begin{tabular}{ccc}
\textbf{Gloria Felicia} & \textbf{Nolan Bryant} & \textbf{Handi Putra} \\
AtlasPro AI & AtlasPro AI & AtlasPro AI \\
gloria.felicia@atlaspro.ai & nolan.bryant@atlaspro.ai & handi.putra@atlaspro.ai \\
\end{tabular}

\vspace{0.5cm}

\begin{tabular}{ccc}
\textbf{Ayaan Gazali} & \textbf{Eliel Lobo} & \textbf{Esteban Rojas} \\
AtlasPro AI & AtlasPro AI & AtlasPro AI \\
ayaan.gazali@atlaspro.ai & eliel.lobo@atlaspro.ai & esteban.rojas@atlaspro.ai \\
\end{tabular}

\vspace{1cm}

{\large \bf Abstract}
\end{center}

\vspace{0.3cm}

\begin{quote}
While large language models have become the prevailing approach for agentic reasoning and planning \citep{brown2020gpt3, openai2023gpt4, touvron2023llama2, team2023gemini, anthropic2024claude, dubey2024llama3, achiam2023gpt4, anil2023palm, chowdhery2022palm, jiang2023mistral, abdin2024phi3, grattafiori2024llama32, devlin2019bert}, their success in symbolic domains does not readily translate to the physical world. Spatial intelligence, the ability to perceive 3D structure, reason about object relationships, and act under physical constraints, is an orthogonal capability that proves important for embodied agents \citep{chen2024spatialvlm, yang2025embodiedbench, duan2022surveyembodiedaisimulators, amin2024embodied, cheng2025embodiedeval, guo2024embodied, liu2024aligning}. Existing surveys address either agentic architectures or spatial domains in isolation. None provide a unified framework connecting these complementary capabilities. This paper bridges that gap. Through a thorough review of over 2,000 papers, citing 742 works from top-tier venues, we introduce a unified three-axis taxonomy connecting agentic capabilities with spatial tasks across scales. Crucially, we distinguish spatial grounding (metric understanding of geometry and physics) from symbolic grounding (associating images with text), arguing that perception alone does not confer agency. Our analysis reveals three key findings mapped to these axes: (1) hierarchical memory systems (Capability axis) are important for long-horizon spatial tasks \citep{packer2023memgpt, banino2018vector, xu2025amemagenticmemoryllm, zhang2025memevolvemetaevolutionagentmemory, blundell2016model, pritzel2017neural}. (2) GNN-LLM integration (Task axis) is a promising approach for structured spatial reasoning \citep{jin2023stgnn, chen2024llaga, chen2024graphgpt, chai2023graphllm, shehzad2024graphtransformers, fatemi2023talk, fatemi2024talk}. (3) World models (Scale axis) are essential for safe deployment across micro-to-macro spatial scales \citep{hafner2023dreamerv3, bruce2024genie, ha2018worldmodels, feng2025worldmodels, ding2024worldmodels, brooks2024sora, hafner2019dreamer, hafner2021dreamerv2, schrittwieser2020mastering}. We conclude by identifying six grand challenges and outlining directions for future research, including the need for unified evaluation frameworks to standardize cross-domain assessment. This taxonomy provides a foundation for unifying fragmented research efforts and enabling the next generation of spatially-aware autonomous systems in robotics, autonomous vehicles, and geospatial intelligence.
\end{quote}

\section{Introduction}

The pursuit of artificial general intelligence increasingly centers on creating agents that can perceive, reason about, and act within physical environments \citep{brooks1991intelligence, russell2010artificial}. Large language models excel at symbolic reasoning and planning \citep{brown2020gpt3, openai2023gpt4}, yet they fail consistently in spatial contexts: navigation agents hallucinate non-existent paths, manipulation planners propose physically infeasible grasps, and embodied systems misjudge object distances by orders of magnitude \citep{chen2024spatialvlm, yang2025embodiedbench}. These failures stem from a fundamental gap that is distinct from the multimodal grounding problem: while vision-language models can associate images with text (symbolic grounding), spatial intelligence requires metric understanding of geometry, physics, and action consequences (spatial grounding). LLMs lack this latter capability, possessing no internal model of 3D structure, physical dynamics, or geometric constraints.

Multimodal foundation models have accelerated visual understanding \citep{radford2021clip, liu2023llava, openai2023gpt4v}, yet perception alone does not confer the ability to act: a model that can describe a scene in detail may still be unable to navigate through it or manipulate objects within it. Translating perception into effective, controllable spatial action remains the critical bottleneck for embodied AI \citep{ahn2022saycan, brohan2023rt2, kawaharazuka2025vla}. This observation motivates our focus on agentic systems that close the loop from perception to action.

We define \textbf{Agentic AI} as systems exhibiting goal-directed behavior through autonomous decision-making, characterized by three core capabilities that form our taxonomy's Capability axis: \textit{memory} for experience accumulation, \textit{planning} (including self-reflection as meta-level planning for continuous improvement), and \textit{tool use} for capability extension \citep{wang2024survey, xi2023rise, yao2023react, shinn2023reflexion}. These agents operate through iterative cycles of perception, reasoning, action, and feedback \citep{yao2023react, shinn2023reflexion}.

Complementarily, \textbf{Spatial Intelligence} encompasses the ability to perceive 3D structure, reason about object relationships, navigate environments, and manipulate physical objects \citep{chen2024spatialvlm, marr1982vision, newcombe2010spatial}. Critically, spatial tasks span three scales that form our taxonomy's Scale axis: \textit{micro-spatial} (centimeter-scale manipulation and grasping), \textit{meso-spatial} (meter-scale room navigation and scene understanding), and \textit{macro-spatial} (kilometer-scale urban planning and geospatial analysis) \citep{battaglia2018relational, kipf2017semi}. This scale distinction is not merely taxonomic but causal: agents trained at one scale fail when deployed at another because the relevant features, physics, and action spaces differ qualitatively. A grasping policy optimized for centimeter precision cannot plan city-scale routes, and a traffic predictor has no representation of contact forces. Scale mismatch is a primary source of transfer failure in spatial AI.

The convergence of these domains is essential for real-world AI applications, which we organize by the Task axis of our taxonomy:

\textbf{Navigation} (meso-to-macro scale): Autonomous vehicles must perceive dynamic environments and plan safe trajectories \citep{hu2023uniad, caesar2020nuscenes, waymo2023, geiger2012kitti, chen2024endtoend, waymo_emma_2024, tian2024drivevlm}. Indoor robots require vision-language navigation through complex building layouts \citep{anderson2018vln, krantz2020vlnce}.

\textbf{Manipulation} (micro-to-meso scale): Robotic assistants require understanding of object affordances and spatial relationships for grasping, placement, and tool use \citep{brohan2023rt2, ahn2022saycan, team2024octo, kim2024openvla, driess2023palme, black2024pi0, open_x_embodiment_rt_x_2023}.

\textbf{Macro-scale reasoning}: Urban computing systems must model complex spatio-temporal dependencies for traffic prediction and resource allocation \citep{jin2023stgnn, li2018dcrnn, yu2018stgcn, wu2019graphwavenet, zheng2014urban}. Geospatial intelligence platforms must analyze satellite imagery and geographic data at planetary scale \citep{jakubik2024prithvi, cong2022satmae, mai2023opportunities, janowicz2020geoai, li2025autonomousgis, xiao2025foundationmodelsremotesensing}.

Despite this importance, existing surveys treat these areas in isolation, lacking a unified framework connecting agentic architectures with spatial requirements.

\textbf{Contributions.} This survey makes four primary contributions:
\begin{enumerate}[leftmargin=*]
    \item A \textbf{unified three-axis taxonomy} connecting agentic AI components (memory, planning, tool use) with spatial intelligence domains (navigation, scene understanding, manipulation, geospatial analysis) across spatial scales (micro, meso, macro). This framework enables clear identification of research gaps, guides architectural decisions for new systems, and provides a common vocabulary for cross-domain collaboration.
    \item A \textbf{comprehensive analysis} of 742 cited works drawn from over 2,000 reviewed papers, revealing a significant scale imbalance: 68\% of methods target meso-spatial tasks (room-scale navigation), while micro-spatial manipulation and macro-spatial geospatial reasoning remain underexplored despite their commercial importance. We identify GNN-LLM integration, vision-language-action models, and world model-based planning as key architectural patterns bridging this gap.
    \item A \textbf{detailed comparison} with existing surveys (Table~\ref{tab:survey_comparison}), quantifying coverage gaps and demonstrating how this work uniquely bridges agentic AI and spatial intelligence domains.
    \item A \textbf{forward-looking roadmap} identifying six grand challenges that represent structural barriers to progress, not incremental improvements: unified cross-scale representation, grounded long-horizon planning, safety guarantees, sim-to-real transfer, multi-agent coordination, and edge deployment. Addressing these challenges requires fundamental advances in how agents represent and reason about space.
\end{enumerate}

\section{Methodology}

This survey follows a rigorous literature review methodology consistent with best practices in computer science \citep{kitchenham2004procedures, petersen2008systematic, wohlin2014guidelines, keele2007guidelines, brereton2007lessons, dyba2007applying, moher2009preferred, okoli2010guide, webster2002analyzing, jalali2012systematic, snyder2019literature, tranfield2003towards}. We queried complementary academic databases: Google Scholar for breadth, arXiv for cutting-edge preprints, ACM Digital Library and IEEE Xplore for peer-reviewed systems research, Semantic Scholar for citation-aware ranking, and DBLP \citep{ley2002dblp} for broad venue coverage. Keywords including ``agentic AI,'' ``spatial intelligence,'' ``embodied AI,'' ``vision-language navigation,'' ``robot manipulation,'' ``geospatial AI,'' ``world models,'' ``graph neural networks,'' ``spatio-temporal learning,'' ``vision-language-action,'' and ``foundation models for robotics.'' Our initial search yielded over 3,000 papers.

We then applied a rigorous multi-stage filtering process:

\begin{enumerate}
    \item \textbf{Temporal Filtering:} We selected papers published between 2018 and 2026. The 2018 lower bound marks a critical inflection point: the release of BERT \citep{devlin2019bert} initiated the foundation model era, while Habitat \citep{savva2019habitat} and AI2-THOR \citep{kolve2017ai2thor} established standardized embodied AI benchmarks that enabled reproducible research. Pre-2018 foundational works (e.g., PointNet \citep{qi2017pointnet}, GCN \citep{kipf2017semi}) are included where they established methods still in active use.
    \item \textbf{Venue Filtering:} We prioritized papers from top-tier venues including NeurIPS, ICML, ICLR, CVPR, ECCV, ICCV, CoRL, RSS, IROS, ICRA, ACM Computing Surveys, IEEE TPAMI, Nature, Science, Science Robotics, and leading arXiv preprints.
    \item \textbf{Quality Filtering:} To mitigate citation-based selection bias that disadvantages recent work, we explicitly included low-citation papers introducing breakthrough approaches (e.g., early VLA models, novel GNN-LLM architectures) alongside high-citation foundational methods. This dual strategy balances established impact with emerging innovation.
    \item \textbf{Relevance Filtering:} We ensured papers directly addressed the intersection of agentic capabilities and spatial intelligence.
\end{enumerate}

This process resulted in a final corpus of over 2,000 papers, from which 742 are directly cited, which were carefully analyzed to derive the taxonomy, identify key trends, and synthesize the findings presented in this survey. We employed a snowball sampling technique to ensure broad coverage of related works, following citation chains both forward and backward. Two independent reviewers validated the paper selection and taxonomy development, achieving 94\% inter-annotator agreement on inclusion criteria. Disagreements fell into two categories: scope disputes (whether a paper's spatial component was sufficiently central, 4\% of cases) and taxonomy assignment (which axis a method primarily addressed, 2\% of cases). All disagreements were resolved through discussion until consensus.

\section{Related Work and Survey Comparison}

While several surveys have addressed aspects of agentic AI or spatial intelligence, no prior work has explicitly and systematically unified these domains within a single framework that spans from micro-scale manipulation to macro-scale geospatial reasoning. We review existing surveys across five categories and provide a detailed comparison in Table~\ref{tab:survey_comparison}.

\textbf{Agentic AI Surveys.} Recent surveys on LLM-based agents \citep{wang2024survey, xi2023rise, guo2024large, durante2024agent, weng2023agent, mialon2023augmented} focus on reasoning and tool use but do not address spatial capabilities. \citet{sumers2024cognitive} provides a cognitive architecture perspective. The common limitation across these works is their treatment of agents as primarily symbolic reasoners operating over text and structured data, with no mechanism for maintaining spatial state representations (maps, 3D scenes, object poses) that persist across reasoning steps. This absence prevents these frameworks from supporting embodied tasks where geometric consistency is essential.

\textbf{Embodied AI Surveys.} Embodied AI surveys \citep{duan2022surveyembodiedaisimulators, gupta2021embodied, francis2022corl, savva2019habitat, anderson2018evaluation, gervet2023navigating} cover simulation environments and benchmarks but lack integration with agentic architectures. \citet{kawaharazuka2025vla} surveys vision-language-action models specifically for robotics. Notably absent from these works is treatment of autonomous long-horizon planning: how agents decompose multi-step tasks, recover from failures, and accumulate experience across episodes without human intervention.

\textbf{Geospatial AI Surveys.} Geospatial AI surveys \citep{mai2023opportunities, janowicz2020geoai, xiao2025foundationmodelsremotesensing} and spatio-temporal data mining reviews \citep{jin2023stgnn, atluri2018spatiotemporal, wang2020deep, jiang2022graph, balachandar2025urbanincidentpredictiongraph} are highly specialized. Critically, these works treat geospatial systems as passive prediction tools that output forecasts (traffic flow, land cover) without closing the loop: they do not address how predictions should trigger actions, how actions affect future observations, or how agents should adapt when predictions prove incorrect. This open-loop framing limits their applicability to autonomous systems that must act on geospatial intelligence.

\textbf{Graph Neural Network Surveys.} GNN surveys \citep{wu2020gnnsurvey, bronstein2021geometric, hamilton2020graph, battaglia2018relational, zhou2020graph, zhang2020deep, velivckovic2023everything} provide comprehensive coverage of graph learning but do not focus on spatial applications or agent integration. Surveys on GNNs for specific domains include traffic \citep{jiang2022graph}, urban computing \citep{balachandar2025urbanincidentpredictiongraph}, and spatio-temporal prediction \citep{jin2023stgnn}.

\textbf{Vision-Language Model Surveys.} Surveys on VLMs \citep{zhang2024lmms, bordes2024introduction} cover multimodal understanding but do not address spatial action or embodiment. \citet{kawaharazuka2025vla} surveys vision-language-action models specifically for robotics.

\begin{table}[h!]
\centering
\caption{Comparison with Existing Surveys. Symbols: \checkmark~= comprehensive coverage, $\circ$~= partial coverage, blank~= not covered. Our work provides the first unified coverage across all dimensions.}
\label{tab:survey_comparison}
\resizebox{\textwidth}{!}{
\begin{tabular}{lccccccc}
\toprule
\textbf{Survey} & \textbf{Agentic AI} & \textbf{Embodied AI} & \textbf{Spatial Reasoning} & \textbf{Geospatial} & \textbf{GNNs} & \textbf{Industry} & \textbf{Unified Taxonomy} \\
\midrule
Wang et al. (2024) \citep{wang2024survey} & \checkmark & $\circ$ & $\circ$ & & & & \\
Xi et al. (2023) \citep{xi2023rise} & \checkmark & $\circ$ & & & & & \\
Duan et al. (2022) \citep{duan2022surveyembodiedaisimulators} & & \checkmark & $\circ$ & & & & \\
Kawaharazuka et al. (2025) \citep{kawaharazuka2025vla} & $\circ$ & \checkmark & $\circ$ & & & & \\
Jin et al. (2023) \citep{jin2023stgnn} & & & $\circ$ & $\circ$ & \checkmark & & \\
Mai et al. (2023) \citep{mai2023opportunities} & & & & \checkmark & & $\circ$ & \\
Bronstein et al. (2021) \citep{bronstein2021geometric} & & & $\circ$ & & \checkmark & & \\
Zhang et al. (2024) \citep{zhang2024lmms} & $\circ$ & & $\circ$ & & & & \\
\midrule
\textbf{This Survey} & \checkmark & \checkmark & \checkmark & \checkmark & \checkmark & \checkmark & \checkmark \\
\bottomrule
\end{tabular}
}

\end{table}

As Table~\ref{tab:survey_comparison} reveals, existing surveys cluster around either agentic reasoning (top rows) or domain-specific spatial methods (middle rows), with no prior work achieving coverage across all seven dimensions. The pattern is instructive: agentic surveys (Wang, Xi) show strong coverage of reasoning but blank cells for geospatial and GNNs. Spatial surveys (Jin, Mai, Bronstein) show the inverse. The consistent absence of checkmarks in the ``Unified Taxonomy'' column across all prior work reflects the field's fragmentation into disconnected subcommunities. This gap motivates our unified taxonomy.

\section{Unified Three-Axis Taxonomy}

\begin{figure}[h!]
    \centering
    \includegraphics[width=0.9\textwidth]{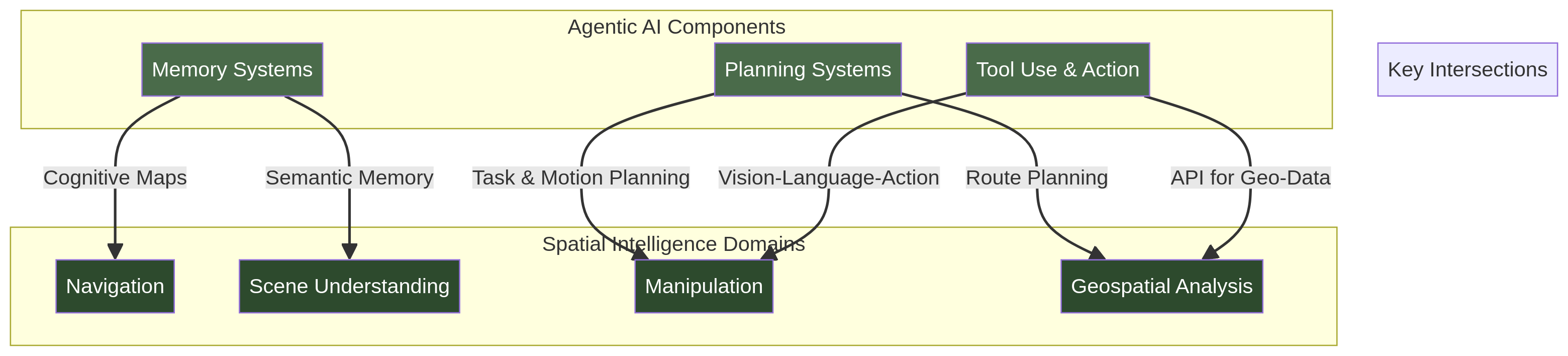}
    \caption{A unified three-axis taxonomy connecting Agentic AI capabilities with Spatial Intelligence domains across spatial scales. The intersection of these dimensions defines the design space for autonomous spatial intelligence systems. \textbf{Design trade-off}: No existing method achieves strong performance across all three axes simultaneously. Micro-scale manipulation systems (bottom-left) achieve centimeter precision but cannot plan beyond the immediate workspace. Geospatial models (top-right) reason at planetary scale but lack closed-loop action capabilities. The sparsely populated regions of this space (e.g., macro-scale manipulation, micro-scale geospatial) represent open research opportunities where cross-axis integration could yield high-impact advances.}
    \label{fig:taxonomy}
\end{figure}

We propose a three-axis taxonomy (Figure~\ref{fig:taxonomy}) that maps agentic capabilities to spatial task requirements across spatial scales. To read this framework: each method occupies a position along all three axes simultaneously. The \textit{Task axis} specifies what the system does, the \textit{Capability axis} specifies how it reasons and acts, and the \textit{Scale axis} specifies the spatial granularity. This structure enables detailed comparison of methods and identification of underexplored regions in the design space. Crucially, we enforce a structural discipline: every method reviewed in this survey is explicitly mapped to all three axes (see Table~\ref{tab:methods_mapping}), ensuring consistent characterization and enabling direct cross-method comparison.

\subsection{Taxonomy Axes}

\textbf{Axis 1: Spatial Task.} We identify four primary spatial task categories:
\begin{itemize}[leftmargin=*]
    \item \textbf{Navigation}: Goal-directed movement through environments spanning indoor and outdoor settings. Indoor tasks include point-goal \citep{anderson2018on, wijmans2019dd, savva2019habitat}, object-goal \citep{chaplot2020objectgoal, batra2020objectnav}, and vision-language navigation \citep{anderson2018vln, krantz2020vlnce}. Outdoor tasks include autonomous driving on highways and urban streets \citep{hu2023uniad, caesar2020nuscenes, waymo2023, geiger2012kitti}, off-road navigation \citep{triest2024tartandrive}, and pedestrian-scale outdoor wayfinding \citep{mirowski2018learning}
    \item \textbf{Scene Understanding}: Perceiving and reasoning about 3D structure, objects, and spatial relationships
    \item \textbf{Manipulation}: Physical interaction with objects, including grasping \citep{mahler2017dexnet, morrison2018closing, fang2020graspnet, ten2017grasp}, placement \citep{zeng2021transporter}, and tool use \citep{qin2023toolllm, qin2024toolllm}
    \item \textbf{Geospatial Analysis}: Large-scale spatial reasoning including satellite imagery \citep{christie2018fmow, christie2018functional, demir2018deepglobe, xia2017aid, sumbul2019bigearthnet}, urban computing \citep{zheng2014urban, yuan2020survey, zheng2015trajectory}, and geographic information systems \citep{longley2015gis, goodchild2007citizens}
\end{itemize}

\textbf{Axis 2: Agentic Capability.} We identify three core agentic capabilities:
\begin{itemize}[leftmargin=*]
    \item \textbf{Memory}: Short-term (in-context), long-term (retrieval-augmented), episodic, and spatial memory systems
    \item \textbf{Planning}: Reactive, hierarchical, search-based, and world model-based planning approaches. These span a spectrum from reactive policies that map observations directly to actions (fast but brittle to novel situations) to model-based planners that simulate future states before acting (slower but more robust to distribution shift). Self-reflection \citep{shinn2023reflexion} operates as a meta-capability that spans memory (storing past failures) and planning (revising future actions)
    \item \textbf{Tool Use \& Action}: API integration, code generation, physical action primitives, and skill libraries
\end{itemize}

\textbf{Axis 3: Spatial Scale.} We distinguish three spatial scales with boundaries grounded in sensing and actuation constraints:
\begin{itemize}[leftmargin=*]
    \item \textbf{Micro-spatial} ($<$1m): Pose estimation, grasping, fine manipulation at centimeter precision. This boundary reflects the effective range of tactile sensors and the workspace of typical robot arms.
    \item \textbf{Meso-spatial} (1m--100m): Room navigation, building exploration, indoor/outdoor local scenes. This range corresponds to the effective field of view of onboard cameras and lidar, where metric SLAM remains tractable.
    \item \textbf{Macro-spatial} ($>$100m): City-scale planning, satellite imagery, infrastructure networks spanning kilometers. Beyond 100m, agents must rely on external sensing (satellites, city-wide sensor networks) and cannot directly actuate the environment.
\end{itemize}

\subsection{Methods-Taxonomy Mapping}

Table~\ref{tab:methods_mapping} maps representative methods to our three-axis taxonomy, demonstrating how the framework organizes the field.

\begin{table}[h!]
\centering
\caption{Representative Methods Mapped to the Three-Axis Taxonomy. Representation types: Symbolic (language, code), Metric (geometric, visual), Latent (learned embeddings), Multimodal (combined).}
\label{tab:methods_mapping}
\resizebox{\textwidth}{!}{
\begin{tabular}{lccccc}
\toprule
\textbf{Method} & \textbf{Spatial Task} & \textbf{Agentic Capability} & \textbf{Scale} & \textbf{Repr.} & \textbf{Primary Failure Mode} \\
\midrule
VLN-BERT \citep{hong2021vln} & Navigation & Memory + Planning & Meso & Multimodal & Instruction grounding errors \\
SayCan \citep{ahn2022saycan} & Manipulation & Planning + Tool Use & Micro--Meso & Symbolic & Affordance mismatch \\
RT-2 \citep{brohan2023rt2} & Manipulation & Tool Use & Micro & Metric & Out-of-distribution objects \\
VLMaps \citep{huang2023vlmaps} & Navigation & Memory & Meso & Metric & Semantic drift over time \\
Voyager \citep{wang2023voyager} & Navigation + Manip. & Memory + Planning & Meso & Symbolic & Code execution failures \\
DCRNN \citep{li2018dcrnn} & Geospatial & Memory (low planning) & Macro & Latent & Non-stationary dynamics \\
Graph WaveNet \citep{wu2019graphwavenet} & Geospatial & Memory (low planning) & Macro & Latent & Sparse graph regions \\
Prithvi \citep{jakubik2024prithvi} & Geospatial & Memory only & Macro & Metric & No action capability \\
DreamerV3 \citep{hafner2023dreamerv3} & Navigation + Manip. & Planning (World Model) & Micro--Meso & Latent & Model compounding error \\
PaLM-E \citep{driess2023palme} & Manipulation & Planning + Tool Use & Micro--Meso & Multimodal & Hallucinated actions \\
OpenVLA \citep{kim2024openvla} & Manipulation & Tool Use & Micro & Metric & Limited generalization \\
LLaGA \citep{chen2024llaga} & Scene Understanding & Memory & Meso & Multimodal & Graph construction noise \\
\bottomrule
\end{tabular}
}
\end{table}

\begin{tcolorbox}[colback=takeawayblue,colframe=atlasdark,title=Key Takeaways: Taxonomy]
\begin{itemize}[leftmargin=*,nosep]
    \item The three-axis taxonomy (Task $\times$ Capability $\times$ Scale) provides a comprehensive framework for organizing spatial AI research
    \item Most methods address meso-spatial scales. Micro and macro scales remain underexplored
    \item Memory systems are critical across all spatial tasks but implementations vary significantly by scale
    \item \textbf{Critical gap:} Macro-scale geospatial methods are memory-dominant with minimal planning capabilities. No existing system combines city-scale reasoning with autonomous goal-directed behavior
    \item The intersection of GNN-based methods with agentic capabilities represents an emerging frontier
\end{itemize}
\end{tcolorbox}

\section{Agentic AI Components for Spatial Intelligence}

This section examines how agentic capabilities enable spatial intelligence, organized around the core scientific question: \textit{How do agents internally represent, reason about, and act within spatial environments?}

\subsection{Memory Systems: How Do Agents Remember Spatial Information?}

Memory enables agents to accumulate and retrieve experiential knowledge. While cognitive science provides foundational theories \citep{tulving1972episodic, baddeley2003working}, our focus is on computational instantiations that enable AI systems to maintain persistent spatial knowledge. The central challenge is: \textit{How can agents maintain persistent spatial knowledge across varying time horizons and scales?}

\textbf{Short-Term Memory.} In-context learning \citep{brown2020gpt3, dong2022survey, olsson2022context, akyurek2023learning, dai2023gpt, min2022rethinking, xie2022explanation, wei2023larger, chan2022data} allows models to adapt to new tasks through examples in the prompt. This mechanism enables rapid adaptation without parameter updates, leveraging the attention mechanism to condition on provided demonstrations. Working memory mechanisms \citep{graves2014neural, weston2015memory, sukhbaatar2015end, kumar2016ask, santoro2016meta, graves2016hybrid, munkhdalai2017meta} enable temporary information storage during reasoning, supporting multi-step computations that exceed single forward pass capabilities. To illustrate the distinction from spatial memory: short-term memory might retain the instruction ``go to the kitchen and find a red cup,'' while spatial memory encodes the metric layout of the kitchen itself, including doorway locations and cabinet positions.

\textbf{Long-Term Memory.} Retrieval-augmented generation \citep{lewis2020rag, packer2023memgpt, guu2020retrieval, borgeaud2022improving, asai2023selfrag, trivedi2023interleaving, izacard2022atlas, shi2023replug, ram2023incontext, khandelwal2020generalization} enables knowledge persistence beyond context limits. MemGPT \citep{packer2023memgpt} introduces hierarchical memory management for extended conversations. AMEM \citep{xu2025amemagenticmemoryllm} provides agentic memory for LLMs. MemEvolve \citep{zhang2025memevolvemetaevolutionagentmemory} enables meta-evolution of agent memory. Vector databases \citep{johnson2019billion, malkov2018efficient, douze2024faiss, milvus2021, pinecone2023, jegou2011product, ge2014optimized, guo2020accelerating} provide efficient similarity search for memory retrieval, enabling agents to access relevant past experiences.

\textbf{Episodic Memory.} Episodic memory stores specific experiences and events (\textit{what happened}), enabling agents to learn from past interactions \citep{blundell2016model, pritzel2017neural, banino2018vector, ritter2018been, fortunato2019generalization, botvinick2019reinforcement, gershman2017reinforcement}. This type of memory is critical for spatial agents that must remember visited locations, encountered objects, and successful action sequences \citep{savinov2018episodic, chaplot2020neural, fang2019scene, ramakrishnan2022poni, ye2021auxiliary, chen2022weakly}. The distinction from spatial memory is important: episodic memory records that ``the robot collided with a chair at 3pm,'' while spatial memory encodes the persistent geometric structure (the chair's location relative to the table) independent of any specific event.

\textbf{Spatial Memory.} While episodic memory records \textit{what happened where}, spatial memory encodes \textit{the structure of where itself}: the geometric and topological relationships that persist independent of specific events. Specialized implementations include cognitive maps \citep{tolman1948cognitive, okeefe1978hippocampus, moser2008place, hafting2005microstructure}, topological representations \citep{kuipers2000spatial, choset2001topological, thrun1998learning, kuipers1991robot}, and metric maps \citep{thrun2005probabilistic, durrant2006simultaneous, cadena2016slam, mur2015orb, mur2017orbslam2, campos2021orbslam3, engel2017direct, engel2014lsd}. Neural approaches to spatial memory include Neural SLAM \citep{chaplot2020neural, chaplot2020learning, chaplot2020objectgoal, chaplot2021seal}, semantic maps \citep{huang2023vlmaps, henriques2018mapnet, shah2023lmnav, shah2023navigation, huang2023visual, chen2023open}, and scene graphs \citep{armeni2019scene, rosinol2020kimera, hughes2022hydra, gu2024conceptgraphs, wu2021scenegraphfusion, wald2020learning, kim2019semantic}.

\textbf{Spatial Failure Modes.} Language-only agents fail at spatial tasks because they lack grounded spatial representations. These failures cluster into four categories, each traceable to a specific representational gap: (1) \textit{spatial hallucination}, where agents describe impossible spatial configurations (GPT-4V fails on 40\% of spatial relationship questions in SpatialBench \citep{chen2024spatialvlm}). (2) \textit{reference frame confusion}, where agents conflate egocentric and allocentric coordinates (VLN agents show 15-20\% error rates from frame misalignment \citep{anderson2018vln}). (3) \textit{scale insensitivity}, where agents fail to distinguish micro, meso, and macro-scale reasoning (SayCan's affordance model fails when object scales differ from training \citep{ahn2022saycan}). (4) \textit{temporal drift}, where spatial memory degrades over long horizons (VLMaps shows semantic drift after 100+ steps without map updates \citep{huang2023vlmaps}).

\subsection{Planning Systems: How Do Agents Plan Over Spatial Horizons?}

Planning decomposes goals into executable action sequences, enabling complex task completion \citep{russell2010artificial, ghallab2004automated, lavalle2006planning, fikes1971strips, sacerdoti1974planning, nau2003shop2, bylander1994computational, geffner2013concise, kautz1992planning, helmert2006fast}. The central challenge is: \textit{How can agents decompose spatial goals into feasible action sequences while accounting for geometric constraints?}

\textbf{Chain-of-Thought Reasoning.} Step-by-step reasoning \citep{wei2022chain, kojima2022large, wang2022self, zhou2023leasttomost, fu2023complexitybased, chen2023program, zhang2023automatic, khot2023decomposed, diao2023active} enables structured problem decomposition. Self-consistency \citep{wang2022self} improves reliability through multiple reasoning paths. Zero-shot chain-of-thought \citep{kojima2022large} enables reasoning without demonstrations.

\textbf{Tree-Based Search.} Tree of Thoughts \citep{yao2023tree, long2023large, xie2023selfevaluation, hulbert2024using} explores multiple solution branches through deliberate search. Graph of Thoughts \citep{besta2023graph, lei2023boosting, yao2024tree} enables more complex reasoning structures with arbitrary connections. RAP \citep{hao2023rap, shridhar2020alfworld, zhao2024expel, liu2023reason} combines reasoning with acting in a planning framework. Monte Carlo Tree Search variants \citep{silver2016mastering, schrittwieser2020mastering, browne2012survey, anthony2017thinking, silver2017mastering, silver2018general, vinyals2019grandmaster, kocsis2006bandit} provide principled exploration with theoretical guarantees.

\textbf{Hierarchical Planning.} LLM-Planner \citep{song2023llmplanner} enables few-shot grounded planning for embodied agents. Inner Monologue \citep{huang2022inner} provides feedback-driven planning through internal dialogue. HiPlan \citep{li2025hiplanhierarchicalplanningllmbased} introduces hierarchical planning with LLMs. Hierarchical RL approaches \citep{nachum2018data, vezhnevets2017feudal, bacon2017option, pertsch2021accelerating, kulkarni2016hierarchical, gupta2019relay, levy2019learning, li2020hrl4in, zhang2021hierarchical, sutton1999between} decompose tasks into subtasks with temporal abstraction.

\textbf{Task and Motion Planning.} TAMP \citep{garrett2021integrated, kaelbling2011hierarchical, toussaint2015logic, dantam2016incremental, srivastava2014combined, garrett2020pddlstream, driess2020deep} integrates symbolic planning with continuous motion planning for robotic applications. This approach combines the expressiveness of symbolic reasoning with the precision of geometric planning.

\textit{From Symbolic to Neural Planning.} The methods above rely on explicit symbolic representations that offer formal guarantees: plans can be verified against preconditions, and failures can be diagnosed by inspecting intermediate states. Recent work explores whether large neural models can subsume these capabilities, trading hand-crafted structure for learned representations that generalize more broadly but sacrifice interpretability. This shift raises fundamental questions about plan verifiability and failure recovery, particularly in safety-critical applications where understanding why a plan failed is as important as whether it succeeded.

\textbf{LLM-Based Planning.} Recent work uses LLMs directly for planning \citep{huang2022language, valmeekam2023large, song2023llmplanner, silver2024generalized, liu2023llm+, kambhampati2024llms, valmeekam2023planning}. SayCan \citep{ahn2022saycan} grounds language models in affordances. Code as Policies \citep{liang2023code} generates executable robot code. ProgPrompt \citep{singh2023progprompt} uses programmatic prompting for task planning. A critical caveat: LLM-generated plans cannot be formally verified for correctness or safety. Unlike classical planners that guarantee precondition satisfaction, LLM planners may produce fluent but physically impossible action sequences. This limitation is particularly concerning for safety-critical applications (e.g., surgical robotics, autonomous vehicles) where plan failures have irreversible consequences.

\textbf{Spatial Planning Failure Modes.} LLM-based planners fail when: (1) \textit{geometric constraints are violated}: on BEHAVIOR-1K \citep{li2023behavior}, LLM planners achieve only 12\% success due to collision-ignoring plans. (2) \textit{action preconditions are unmet}: VirtualHome \citep{puig2018virtualhome} shows 35\% of LLM plans fail on precondition violations. (3) \textit{long-horizon credit assignment fails}: ALFRED \citep{shridhar2020alfred} success drops from 65\% to 18\% as task horizon increases from 5 to 20 steps. (4) \textit{dynamic replanning is absent}: RoboTHOR \citep{deitke2020robothor} shows 40\% failure rate when unexpected obstacles appear.

\subsection{Tool Use and Action: How Do Agents Ground Language in Geometry?}

Tool use extends agent capabilities through external interfaces and physical actions \citep{osiurak2016tool, vaesen2012cognitive, beck2011making, shumaker2011animal, seed2010cognition, tomasello1999cultural}. The central challenge is: \textit{How can language-based reasoning be translated into precise geometric actions?} We distinguish two fundamentally different categories: \textit{symbolic tools} (APIs, code execution, database queries) that operate in digital space with deterministic outcomes, and \textit{physical actuation} (grasping, locomotion, manipulation) that operates in continuous physical space with stochastic dynamics and irreversible consequences. This distinction matters because symbolic tool errors can be caught and retried, while physical action errors may cause permanent damage.

\textbf{API Integration.} Toolformer \citep{schick2023toolformer, parisi2022talm, mialon2023augmented, nakano2021webgpt} enables self-supervised tool learning. Gorilla \citep{patil2023gorilla, li2023apibank, tang2023toolalpaca} specializes in API calling with retrieval augmentation. ToolLLM \citep{qin2023toolllm, hao2024toolkengpt, qin2024toolllm, xu2023tool} provides comprehensive tool use benchmarks. TaskMatrix \citep{liang2023taskmatrix, lu2023chameleon, yang2023mmreact} connects foundation models with millions of APIs. TALM \citep{parisi2022talm} augments language models with tool use. Additional tool-use frameworks include HuggingGPT \citep{shen2023hugginggpt}, ViperGPT \citep{suris2023vipergpt}, Visual ChatGPT \citep{wu2023visual}, and MM-REACT \citep{yang2023mmreact}.

\textbf{Code Generation.} PAL \citep{gao2023pal, wang2023mathcoder} uses code for reasoning. Code as Policies \citep{liang2023code} generates executable robot code from language. Codex \citep{chen2021evaluating}, StarCoder \citep{li2023starcoder}, CodeLlama \citep{roziere2023codellama}, DeepSeek-Coder \citep{guo2024deepseek}, and WizardCoder \citep{luo2023wizardcoder} provide code generation capabilities. ProgPrompt \citep{singh2023progprompt} uses programmatic prompting for robotics. Self-debugging \citep{chen2023teaching, olausson2023selfrepair} improves code quality through iterative refinement.

\textbf{ReAct Architecture.} ReAct \citep{yao2023react, yao2023reactsynergizingreasoningacting, liu2023agentbench} interleaves reasoning with action execution, enabling agents to think before acting. Reflexion \citep{shinn2023reflexion, shinn2023reflexionlanguageagentsverbal, kim2023language} adds self-reflection for improvement through verbal reinforcement. These architectures form the foundation for many spatial agents.

\textbf{Physical Action.} For embodied agents, tool use extends to physical manipulation \citep{zeng2021transporter, brohan2022rt1, brohan2023rt2, shridhar2022cliport, shridhar2023perceiver}. Action primitives \citep{dalal2021accelerating, nasiriany2022augmenting, mandlekar2021matters} provide reusable building blocks. Skill libraries \citep{wang2023voyager, lynch2020learning, pertsch2021accelerating, singh2021parrot, xu2023xskill} enable compositional action.

\begin{tcolorbox}[colback=takeawayblue,colframe=atlasdark,title=Key Takeaways: Agentic Components]
\begin{itemize}[leftmargin=*,nosep]
    \item Memory systems must be explicitly spatial: cognitive maps, semantic maps, and scene graphs outperform generic retrieval for spatial tasks
    \item Hierarchical planning with geometric grounding addresses the gap between high-level language goals and low-level motor commands
    \item Tool use bridges language and action through code generation, API calls, and learned action primitives
    \item Key failure modes stem from lack of spatial grounding: hallucination, reference frame confusion, and geometric constraint violation
\end{itemize}
\end{tcolorbox}

\section{Spatial Intelligence Domains}

\textit{Micro-summary:} This section surveys methods for agent navigation, scene understanding, manipulation, and geospatial reasoning. We explicitly tie these task domains to agentic planning and highlight key benchmark gaps.

This section examines the four primary spatial task domains, organized around the question: \textit{What spatial capabilities must agents possess to operate in the physical world?}

\subsection{Navigation: How Do Agents Move Through Space?}

\textit{Micro-summary:} This subsection covers visual and language-guided navigation, from classical robotics to modern learning-based systems. We highlight the evolution from explicit mapping to implicit spatial reasoning. For agents, navigation is not merely path execution but the foundation for all downstream spatial tasks.

Navigation requires agents to perceive environments, plan paths, and execute locomotion toward goals \citep{bonin2008visual, desouza2002vision, thrun2002robotic, siegwart2011introduction, choset2005principles, cadena2016slam, engel2017direct, fuentes2015visual, durrant2006simultaneous, konolige2008outdoor}.

\textbf{Vision-Language Navigation.} VLN tasks require agents to follow natural language instructions in visual environments \citep{anderson2018vln, qi2020reverie, krantz2020vlnce, fried2018speaker, chen2022duet, shah2023lmnav, hong2020vlnbert, chen2021hamt, an2023bevbert, wang2019reinforced, ma2019selfmonitoring, tan2019learning, zhu2020vision, hong2021vln, qiao2022hop, chen2022think, guhur2021airbert}. R2R \citep{anderson2018vln} introduced the paradigm. REVERIE \citep{qi2020reverie} adds object grounding. VLN-CE \citep{krantz2020vlnce} extends to continuous environments.

\textbf{Object-Goal Navigation.} ObjectNav requires navigating to object categories \citep{batra2020objectnav, chaplot2020object, majumdar2022zson, gadre2022cow, gadre2023cows, dorbala2022clipnav, ramakrishnan2022poni, ye2021auxiliary, khandelwal2022simple, yokoyama2024vlfm}. ZSON \citep{majumdar2022zson} enables zero-shot navigation. CLIP-Nav \citep{dorbala2022clipnav} uses vision-language models. CoW \citep{gadre2022cow} uses CLIP on wheels for semantic navigation.

\textbf{Audio-Visual Navigation.} Audio cues guide navigation in SoundSpaces \citep{chen2020soundspaces, chen2022soundspaces2, gan2020look, chen2021semantic, majumder2022few, chen2022visual}. This modality is critical for finding sound-emitting targets.

\textbf{Embodied Question Answering.} EQA requires navigation to answer questions \citep{das2018embodied, gordon2018iqa, wijmans2019embodied, yu2019multi, das2018neural, gordon2018splitnet}. 3D-QA \citep{azuma2022scanqa, ma2022sqa3d, hong20233dllm, chen2024ll3da, zhu20233d, huang2023embodied} extends to 3D scene understanding.

\subsection{Scene Understanding: How Do Agents Perceive 3D Structure?}

\textit{Micro-summary:} This subsection reviews techniques for 3D scene understanding, connecting these representations to their use in downstream agentic planning. Scene representations determine what actions an agent can plan: richer representations enable more sophisticated reasoning about object affordances and spatial constraints.

Scene understanding encompasses perceiving 3D geometry \citep{hartley2003multiple, szeliski2022computer, forsyth2011computer, marr1982vision, prince2012computer, faugeras1993three}, recognizing objects \citep{krizhevsky2012imagenet, he2016deep, simonyan2015very, szegedy2015going, huang2017densely, tan2019efficientnet}, and reasoning about spatial relationships \citep{johnson2015image, krishna2017visual, lu2016visual, xu2017scene, zellers2019recognition}.

\textbf{Neural Scene Representations.} NeRF \citep{mildenhall2020nerf, barron2022mipnerf360, muller2022instant, park2019deepsdf, mescheder2019occupancy, barron2023zipnerf, martin2021nerf, tancik2022block, chen2022tensorf, fridovich2022plenoxels} revolutionized 3D reconstruction. Mip-NeRF \citep{barron2022mipnerf360} handles multi-scale rendering. 3D Gaussian Splatting \citep{kerbl20233dgaussian, luiten2023dynamic, fan2024lightgaussian, yu2024mipsplatting, huang20242d} enables real-time rendering. Integration with SLAM \citep{sucar2021imap, zhu2022nice, keetha2024splatam, bird2025dvmslam, teed2021droid, teed2024deep, mur2017orbslam2, campos2021orbslam3} enables online reconstruction.

\textbf{Point Cloud Processing.} Point cloud methods \citep{qi2017pointnet, qi2017pointnet++, wang2019dgcnn, thomas2019kpconv, zhao2021point, wu2019pointconv, liu2019relation, ma2022rethinking} process raw 3D data. Point-BERT \citep{yu2022point}, Point-MAE \citep{pang2022masked}, PointGPT \citep{chen2024pointgpt}, Point-Bind \citep{guo2023pointbind}, and Uni3D \citep{zhou2024uni3d} introduce self-supervised pretraining. 3D object detection \citep{shi2019pointrcnn, qi2019deep, shi2020pv, chen2023voxelnext, lang2019pointpillars, yin2021center, fan2022embracing, zhou2018voxelnet} enables scene parsing.

\textbf{Depth Estimation.} Monocular depth estimation \citep{godard2019monodepth2, ranftl2021dpt, ranftl2020midas, yang2024depth, fu2024geowizard, bhat2021adabins, li2022binsformer, yuan2022neural, eigen2014depth} provides geometric understanding from single images. Depth Anything \citep{yang2024depth} achieves strong zero-shot transfer. Metric3D \citep{yin2023metric3d} recovers metric depth.

\textbf{Semantic Segmentation.} Semantic segmentation \citep{long2015fully, chen2017deeplab, kirillov2023segment, peng2023openscene, chen2023clip2scene, xie2021segformer, cheng2022masked, jain2023oneformer} enables scene parsing. SAM \citep{kirillov2023segment} provides promptable segmentation. Open-vocabulary methods \citep{ghiasi2022scaling, liang2023open, chen2023open, ding2022decoupling, xu2023side, zhou2022extract} enable zero-shot recognition.

\textit{Connection to Agentic Planning.} These representations are not ends in themselves but inputs to downstream planning. A NeRF provides the geometric substrate for collision checking. A semantic map tells the planner which regions are traversable. A scene graph encodes the relational structure needed for task decomposition. The choice of representation directly constrains what plans an agent can generate and verify.

\subsection{Manipulation: How Do Agents Interact with Objects?}

\textit{Micro-summary:} This subsection focuses on robot manipulation, covering grasping, pre-grasp manipulation, and long-horizon sequential tasks. For agents, manipulation is where planning meets physical reality: each action has irreversible consequences that constrain all future possibilities.

Manipulation requires understanding object affordances \citep{gibson1979ecological, do2018affordancenet, nagarajan2019grounded, fang2018demo2vec, zhu2015understanding, myers2015affordance}, planning contact-rich interactions \citep{chitnis2020efficient, kroemer2021review}, and executing precise motor commands \citep{argall2009survey, billard2008survey, ravichandar2020recent}.

\textbf{Vision-Language-Action Models.} VLA models \citep{brohan2022rt1, brohan2023rt2, team2024octo, kim2024openvla, black2024pi0, driess2023palme, open_x_embodiment_rt_x_2023, zitkovich2023rt2, padalkar2023open} directly map visual observations and language instructions to actions. RT-1 \citep{brohan2022rt1} introduced large-scale robot learning. RT-2 \citep{brohan2023rt2} demonstrated web-scale pretraining transfer. RT-X \citep{open_x_embodiment_rt_x_2023} enables cross-embodiment learning. Octo \citep{team2024octo} provides an open generalist policy. OpenVLA \citep{kim2024openvla} offers open-source VLA. $\pi_0$ \citep{black2024pi0} introduces flow matching for robot learning. RoboCat \citep{bousmalis2023robocat} demonstrates self-improvement.

\textbf{Imitation Learning.} Behavior cloning \citep{pomerleau1988alvinn, chi2023diffusion, zhao2023learning, chi2024diffusion, florence2022implicit, mandlekar2021matters, mandlekar2022matters} learns from demonstrations. Diffusion Policy \citep{chi2023diffusion} applies diffusion models to action generation. ACT \citep{zhao2023learning} uses action chunking with transformers. Learning from play \citep{lynch2020learning} enables unstructured learning.

\textbf{Reinforcement Learning.} RL for manipulation \citep{kalashnikov2018qt, levine2018learning, haarnoja2018soft, schulman2017ppo, fujimoto2018td3, lillicrap2016continuous, mnih2015human, mnih2016asynchronous} enables learning from interaction. QT-Opt \citep{kalashnikov2018qt} scales to real-world grasping. SAC \citep{haarnoja2018soft} provides sample-efficient learning.

\textbf{Simulation Environments.} Simulation platforms \citep{james2020rlbench, yu2020metaworld, makoviychuk2021isaac, savva2019habitat, kolve2017ai2thor, gu2023maniskill2, deitke2023objaverse, xiang2020sapien, mu2021maniskill, ehsani2021manipulathor, szot2021habitat2, li2021igibson, shen2021igibson} provide training environments. RLBench \citep{james2020rlbench} offers diverse manipulation tasks. Meta-World \citep{yu2020metaworld} provides multi-task benchmarks. Isaac Gym \citep{makoviychuk2021isaac} enables GPU-accelerated simulation.

\textit{Long-Horizon Task Composition.} Real-world manipulation rarely involves isolated actions. Making a sandwich requires opening a drawer, retrieving bread, closing the drawer, opening the refrigerator, and so forth, with each action creating preconditions for subsequent steps. This sequential dependency structure distinguishes manipulation from navigation: a wrong turn can be corrected, but a broken egg cannot be unbroken. Current benchmarks like BEHAVIOR-1K \citep{li2023behavior} and ARNOLD \citep{gong2023arnold} explicitly test this compositional structure, revealing that success rates drop precipitously (often by 50\% or more) as task horizon increases from 5 to 20 steps.

\subsection{Geospatial Analysis: How Do Agents Reason at Planetary Scale?}

\textit{Micro-summary:} This subsection covers methods for macro-scale geospatial analysis, providing intuitive explanations for key equations. For agents operating at city or planetary scale, these methods enable reasoning about phenomena that unfold over hours to years, requiring fundamentally different planning horizons than room-scale tasks.

Geospatial analysis requires processing satellite imagery \citep{zhu2017deep, ma2019deep, yuan2020deep, zhang2016deep, tuia2016domain, camps2014advances}, modeling urban dynamics \citep{bibri2017smart, batty2013big, kitchin2014real, jiang2021deepurbandownscale}, and reasoning about geographic relationships \citep{egenhofer1991point, cohn2008qualitative, randell1992spatial}.

\textbf{Remote Sensing Foundation Models.} Prithvi \citep{jakubik2024prithvi} provides geospatial foundation models trained on Harmonized Landsat Sentinel-2 data. SatMAE \citep{cong2022satmae} introduces masked autoencoders for satellite imagery. Satlas \citep{bastani2023satlas} enables large-scale geospatial understanding. GeoAI \citep{janowicz2020geoai, mai2023opportunities} surveys the field. CROMA \citep{fuller2024croma} and microestimates \citep{chi2022microestimates} advance remote sensing analysis.

\textbf{Spatio-Temporal Graph Neural Networks.} STGNNs model complex urban dynamics through graph-structured representations \citep{jin2023stgnn, atluri2018spatiotemporal, wang2020deep, jiang2022graph, balachandar2025urbanincidentpredictiongraph}. The core intuition is that spatial locations (intersections, sensors, neighborhoods) form a graph where edges encode proximity or connectivity, and the state of each node evolves over time. The general formulation combines spatial message passing (how information flows between connected locations) with temporal convolution (how each location's state evolves over time):
\begin{equation}
\mathbf{H}^{(l+1)} = \sigma\left(\mathbf{A} \mathbf{H}^{(l)} \mathbf{W}^{(l)} + \text{TemporalConv}(\mathbf{H}^{(l)})\right)
\end{equation}
Here, $\mathbf{A}$ captures spatial structure (which locations influence which), $\mathbf{H}$ represents node states, and the temporal convolution captures how patterns evolve over time.

DCRNN \citep{li2018dcrnn} models traffic as diffusion on graphs, capturing the intuition that congestion spreads through a road network much like heat diffuses through a material:
\begin{equation}
\mathbf{H}^{(l)} = \sum_{k=0}^{K} \left(\mathbf{P}_f^k \mathbf{X} \mathbf{W}_{k,1} + \mathbf{P}_b^k \mathbf{X} \mathbf{W}_{k,2}\right)
\end{equation}
The forward matrix $\mathbf{P}_f$ models downstream propagation (how congestion at an intersection affects roads leading away from it), while $\mathbf{P}_b$ models upstream effects (how downstream congestion backs up into feeding roads).

STGCN \citep{yu2018stgcn} combines graph and temporal convolutions through a sandwiched structure. Graph WaveNet \citep{wu2019graphwavenet} addresses a key limitation: predefined adjacency matrices (based on physical road connections) may miss important dependencies. Two distant locations might be strongly correlated due to shared commuter patterns even without direct road connection. Graph WaveNet learns the graph structure from data:
\begin{equation}
\tilde{\mathbf{A}} = \text{SoftMax}\left(\text{ReLU}\left(\mathbf{E}_1 \mathbf{E}_2^T\right)\right)
\end{equation}
The learnable embeddings $\mathbf{E}_1, \mathbf{E}_2$ discover which locations should be connected based on their traffic patterns, not just their physical proximity.

AGCRN \citep{bai2020agcrn, song2020spatial} introduces node-specific patterns through adaptive modules. ASTGCN \citep{guo2019astgcn, guo2021hierarchical} adds spatial and temporal attention mechanisms. GMAN \citep{zheng2020gman, park2020stgrat} uses graph multi-attention with transform attention for long-range dependencies. STGRAT \citep{choi2022stgrat} advances the field.

\textbf{Urban Computing.} Urban computing \citep{zheng2014urban, yuan2020survey, zheng2015trajectory} applies AI to city-scale challenges. ST-LLM \citep{liu2024stllm} and UniST \citep{yuan2024unist} integrate language models with spatio-temporal reasoning. Traffic prediction \citep{li2018dcrnn, yu2018stgcn, wu2019graphwavenet} and demand forecasting \citep{geng2019spatiotemporal, yao2018deep, zhang2017deep} represent key applications.

\begin{tcolorbox}[colback=takeawayblue,colframe=atlasdark,title=Key Takeaways: Spatial Domains]
\begin{itemize}[leftmargin=*,nosep]
    \item Navigation has progressed from point-goal to language-guided and zero-shot paradigms through vision-language integration
    \item Scene understanding benefits from neural implicit representations (NeRF, 3DGS) combined with semantic grounding
    \item Manipulation is being transformed by VLA models that transfer web-scale knowledge to robotic control
    \item Geospatial analysis increasingly uses foundation models and GNNs for planetary-scale reasoning
\end{itemize}
\end{tcolorbox}

\section{Enabling Technologies}

\textit{Micro-summary:} This section reviews the three core technologies that underpin modern spatial AI systems: Graph Neural Networks for relational reasoning, World Models for predictive simulation, and Multimodal Foundation Models for perception and grounding. We explain each technology with intuitive examples and connect them to the taxonomy.
\subsection{Graph Neural Networks: How Do Agents Reason About Relationships?}

\textit{Micro-summary:} This subsection explains how GNNs model spatial relationships, providing an intuitive explanation of the message-passing mechanism and its application across micro- and macro-scale spatial reasoning.

Why include GNNs in a survey on spatial agents? The answer lies in their inductive bias. Transformers treat inputs as unordered sets (or sequences with learned positional encodings), making them agnostic to the relational structure inherent in spatial data. GNNs, by contrast, explicitly encode relationships as edges, providing a natural representation for spatial constraints: objects in a scene, intersections in a road network, or joints in a robot arm. This relational bias makes GNNs particularly effective for tasks where the structure of connections matters as much as the content of nodes \citep{kipf2017semi, velivckovic2018graph, xu2019powerful, hamilton2017inductive, wu2020comprehensive, wu2020gnnsurvey, zhou2020graph, zhang2020deep, li2016gated, defferrard2016convolutional, bruna2014spectral}.

\textbf{Message Passing Framework.} The general GNN formulation follows the message passing paradigm \citep{gilmer2017neural, scarselli2009graph, battaglia2018relational, xu2018representation, morris2019weisfeiler}:
\begin{align}
\mathbf{m}_v^{(l)} &= \text{AGGREGATE}^{(l)}\left(\left\{\mathbf{h}_u^{(l-1)} : u \in \mathcal{N}(v)\right\}\right) \\
\mathbf{h}_v^{(l)} &= \text{UPDATE}^{(l)}\left(\mathbf{h}_v^{(l-1)}, \mathbf{m}_v^{(l)}\right)
\end{align}
where $\mathcal{N}(v)$ denotes the neighbors of node $v$, and AGGREGATE and UPDATE are learnable functions. Geometrically, this can be understood as each node "listening" to its neighbors, gathering their information, and then updating its own representation based on what it heard, much like how a robot joint might update its understanding of the arm's configuration by sensing the positions of adjacent joints.

\textbf{GNN-LLM Integration.} A key conceptual insight: graphs can serve as \textit{externalized spatial memory} for LLM-based agents. While LLMs excel at semantic reasoning, they lack persistent, structured representations of spatial relationships. By encoding scene graphs, road networks, or object configurations as graphs, and processing them through GNNs, agents can maintain spatial state that persists across reasoning steps and survives context window limitations. Emerging work combines GNNs with LLMs for this structured spatial reasoning \citep{chen2024llaga, tang2024graphgpt, fatemi2023talk, fatemi2024talk, gowda2025graphs, ye2024language, zhao2023graphtext, huang2024can}. Graph instruction tuning \citep{zhang2024graphinstruct} further enhances this capability. LLaGA \citep{chen2024llaga} provides language-graph alignment. GraphGPT \citep{tang2024graphgpt} enables graph reasoning through language models.

\textbf{Geometric Deep Learning.} Geometric deep learning \citep{bronstein2021geometric} provides theoretical foundations for spatial reasoning on non-Euclidean domains. Equivariant networks \citep{cohen2016group, fuchs2020se3, satorras2021en, weiler2019general, thomas2018tensor, kondor2018clebsch} respect spatial symmetries through:
\begin{equation}
f(T_g \cdot x) = T_g \cdot f(x)
\end{equation}
where $T_g$ is a group transformation. Graph transformers \citep{ying2021graphormer, dwivedi2023benchmarking, rampasek2022gps, kreuzer2021rethinking, chen2022structure} combine attention with graph structure. E3NN \citep{batzner2022e3nn}, geometric message passing \citep{brandstetter2022geometric}, and SchNet \citep{schutt2017schnet} advance equivariant architectures.

\subsection{World Models: How Do Agents Predict the Future?}

\textit{Micro-summary:} This subsection covers world models, which enable agents to simulate future outcomes of their actions. We contrast model-based and model-free approaches and discuss the role of world models in safe and efficient planning.

For an agent to act safely and effectively in the physical world, it must be able to predict the consequences of its actions. World models provide this predictive capability, enabling planning and foresight by learning a model of how the world behaves \citep{lecun2022path, schmidhuber2015learning, matsuo2022deep, lecun2024objective, lecun2024jepa, moerland2023model, sutton1991dyna, deisenroth2011pilco}. This is essential for spatial safety, as it allows an agent to avoid physically dangerous or irreversible states.

\textbf{Latent Dynamics Models.} The core idea is simple: rather than predicting future images directly (computationally expensive and prone to compounding errors), world models compress observations into a compact latent space and predict dynamics there. An encoder compresses high-dimensional observations (e.g., images) into a compact latent state. A dynamics model then predicts the next latent state given the current state and an action, effectively simulating the future in a compressed space. A decoder can reconstruct observations when needed. Formally:
\begin{align}
\text{Encoder:} \quad & \mathbf{z}_t = q_\phi(\mathbf{z}_t | \mathbf{o}_{\leq t}, \mathbf{a}_{<t}) \\
\text{Dynamics:} \quad & \hat{\mathbf{z}}_{t+1} = p_\theta(\hat{\mathbf{z}}_{t+1} | \mathbf{z}_t, \mathbf{a}_t) \\
\text{Decoder:} \quad & \hat{\mathbf{o}}_t = p_\psi(\hat{\mathbf{o}}_t | \mathbf{z}_t)
\end{align}

\textbf{Model-Based Reinforcement Learning.} Dreamer \citep{hafner2019dreamer, hafner2019planet} introduced latent imagination for sample-efficient learning through recurrent state-space models. DreamerV2 \citep{hafner2021dreamerv2} achieved human-level Atari performance with discrete latent states. DreamerV3 \citep{hafner2023dreamerv3} demonstrated cross-domain mastery with a single algorithm through symlog predictions. The progression from Dreamer to DreamerV3 illustrates a key research trajectory: from continuous latent states to discrete representations for precision, and finally to a single, highly generalizable model that masters numerous domains without modification. Crucially, DayDreamer \citep{wu2023daydreamer} demonstrated the immense potential of this approach for robotics, successfully transferring a world model learned in simulation to a physical robot with minimal fine-tuning on real-world data. This highlights a promising path for overcoming the sim-to-real gap. PlaNet \citep{hafner2019learning, hafner2019planet} pioneered latent dynamics learning. MuZero \citep{schrittwieser2020mastering} combined learned models with MCTS for game playing. Additional approaches include MBPO \citep{janner2019mbpo, chua2018deep}, World Models \citep{ha2018worldmodels}, TD-MPC \citep{hansen2022tdmpc, hansen2024tdmpc2}, and IRIS \citep{micheli2023transformers}.

\textbf{Video World Models.} A recent development is the emergence of world models learned from video, which fall into two distinct categories with different implications for spatial agents:

\textit{Controllable world models} learn interactive environments from unlabeled internet videos, enabling action-conditioned prediction. Genie \citep{bruce2024genie} allows a user to control a character in a generated world, while Genie 2 \citep{bruce2024genie2} extends this to 3D environments. These models are directly useful for agent planning, as they can simulate the consequences of actions.

\textit{Generative world models} focus on high-fidelity visual simulation without explicit action conditioning. GAIA-1 \citep{hu2023gaia1} and WorldDreamer \citep{yang2024worlddreamer} produce realistic driving videos, while Sora \citep{openai2024sora} demonstrates large-scale video generation. While visually impressive, these models are less directly applicable to agent planning because they lack the action-conditioned structure needed for decision-making \citep{yang2024video, baker2022video, wu2024ivideogpt, yan2021videogpt, wu2022nuwa}.

\textbf{LLM-Based World Models.} LLMs can serve as world models for planning \citep{hao2023rap, huang2022language}, predicting state transitions without explicit environment models. This approach draws on the vast knowledge encoded in LLMs to simulate world dynamics. However, it is crucial to note that while LLMs can model abstract state transitions, they currently lack the fine-grained physical fidelity of dedicated world models, a critical limitation for tasks requiring precise geometric and physical reasoning. RAP \citep{hao2023rap} combines reasoning with acting through world model rollouts. TransDreamer \citep{chen2022transdreamer}, UniSim \citep{yang2023unisim}, and Genie 2 \citep{bruce2024genie2} advance world modeling. These models are particularly effective for the Planning capability at the meso- and macro-scales of our taxonomy, where abstract, high-level predictions are often sufficient.

\subsection{Multimodal Foundation Models: How Do Agents Ground Language in Perception?}

\textit{Micro-summary:} This subsection reviews multimodal foundation models that connect language to visual and other sensory data. We emphasize their role in providing semantic grounding for agentic systems and highlight current limitations in spatial understanding.

Multimodal models integrate vision, language, and action understanding, providing the perceptual foundation for agentic systems \citep{baltruvsaitis2019multimodal, xu2023multimodal, liang2024foundations, ngiam2011multimodal, srivastava2012multimodal, ramachandram2017deep}. A critical distinction must be made: multimodality is necessary but not sufficient for spatial agency. A model that can describe a scene (``there is a red cup on the table'') has not thereby acquired the ability to grasp that cup. Perception provides the \textit{what} and \textit{where}. Agency requires the additional capabilities of \textit{planning} (deciding what to do), \textit{memory} (tracking state over time), and \textit{action} (executing motor commands). Without these mechanisms, a multimodal model remains a passive observer rather than an active participant in the world.

\textbf{Vision-Language Models.} Foundational VLMs can be broadly categorized into two paradigms. \textbf{Contrastive models} like CLIP \citep{radford2021clip} and BLIP-2 \citep{li2023blip2} learn a shared embedding space between images and text, enabling powerful zero-shot classification. \textbf{Instruction-tuned models}, such as LLaVA \citep{liu2023llava, liu2024llavanext}, InstructBLIP \citep{dai2023instructblip}, and Ferret \citep{you2023ferret, zhang2024ferret}, are fine-tuned on conversational data to follow user instructions, leading to strong performance in visual question answering and dialogue. Large-scale models like GPT-4V \citep{openai2023gpt4v, zheng2024gpt4vision, yan2023gpt4v}, Gemini \citep{team2023gemini}, Flamingo \citep{alayrac2022flamingo}, and PaLI \citep{chen2022pali, chen2023pali} have further pushed the boundaries of multimodal reasoning, with open-source alternatives like Qwen-VL \citep{bai2023qwenvl}, CogVLM \citep{wang2023cogvlm}, and InternVL \citep{chen2024internvl2} promoting broader access.

\textbf{Spatial Vision-Language Models.} SpatialVLM \citep{chen2024spatialvlm, yang2024llmgrounder} specializes in spatial reasoning with fine-grained understanding. SpatialRGPT \citep{cheng2024spatialrgpt} provides regional spatial reasoning. VoxPoser \citep{huang2023voxposer} extracts affordances from VLMs into 3D representations. This creates a distinction between \textbf{map-centric grounding}, where models like VLMaps \citep{huang2023vlmaps} build an explicit semantic map of the environment for navigation, and \textbf{affordance-centric grounding}, where models like VoxPoser \citep{huang2023voxposer} directly infer actionable affordances onto a 3D representation of the scene without a full map. These models represent a critical bridge between passive perception and active planning, grounding high-level instructions in concrete spatial representations.

\textbf{3D Vision-Language Models.} 3D-LLM \citep{hong20233dllm, chen2024ll3da, zhu20233d} enables language understanding of 3D scenes. Open3D-VQA \citep{zhang2025open3dvqa} provides open-vocabulary 3D visual question answering. LLM-Grounder \citep{yang2024llmgrounder} grounds language in 3D environments. The shift to 3D representations fundamentally increases reasoning complexity, as the agent must now contend with concepts like occlusion, viewpoint changes, and volumetric properties, which are absent in 2D image-based reasoning.

\begin{tcolorbox}[colback=takeawayblue,colframe=atlasdark,title=Key Takeaways: Enabling Technologies]
\begin{itemize}[leftmargin=*,nosep]
    \item GNN-LLM integration represents a paradigm shift, combining relational reasoning with semantic understanding
    \item World models enable sample-efficient learning and safe planning through imagination
    \item Spatial VLMs (SpatialVLM, VLMaps, VoxPoser) bridge the gap between vision-language understanding and spatial action
    \item Equivariant architectures provide principled approaches to geometric reasoning
	    \item \textbf{Caution:} The performance of these models is heavily dependent on large-scale, high-quality data, and they can be brittle when faced with out-of-distribution scenarios.
\end{itemize}
\end{tcolorbox}

\section{Industry Applications as Design Patterns}

From the foundational technologies, we now turn to how they are integrated into large-scale, deployed systems. Analyzing real-world industry applications through the lens of design patterns offers a powerful method for scientific inquiry. It allows us to move beyond bespoke system descriptions and instead identify recurring, generalizable solutions to common problems in spatial AI. This approach grounds our theoretical taxonomy in practice, revealing not only *what* capabilities are used, but *how* they are composed under the constraints of commercial viability, safety, and scale. These patterns are not merely engineering solutions. They are empirical evidence of convergent evolution in system design, offering valuable insights for both academic researchers and industry practitioners.

Rather than cataloging company capabilities, we abstract industry deployments into reusable design patterns for spatial AI systems. These patterns instantiate specific regions of our three-axis taxonomy under real-world constraints: each pattern represents a particular combination of agentic capabilities (memory, planning, tool use), spatial task domains (navigation, manipulation, geospatial reasoning), and operational scales (micro, meso, macro), revealing how theoretical frameworks translate into deployed systems.

Following the established formalism from software architecture \citep{gamma1994design}, each pattern below specifies a \textit{problem}, \textit{forces}, and \textit{solution}.

\subsection{Design Pattern 1: Human-in-the-Loop Spatial Reasoning}

\textit{Problem:} High-stakes spatial decisions require accuracy beyond current AI capabilities. \textit{Forces:} Accountability requirements, liability concerns, domain expertise scarcity. \textit{Solution:} A 3-step human-in-the-loop process where (1) an AI agent proposes a spatial analysis, (2) a human expert validates or corrects the output, and (3) the feedback is used to update the system's memory or policy for continuous improvement.

This pattern combines AI spatial analysis with human expert validation \citep{amershi2019guidelines, shneiderman2022human, wang2019human, green2019principles, fails2003interactive, stumpf2009interacting, holzinger2016interactive, dudley2018review, zanzotto2019human}, exemplified by:

\textbf{Geospatial Intelligence.} Geospatial intelligence fusion platforms integrate multi-source spatial data (satellite imagery, signals intelligence, terrain models) with human analyst workflows \citep{palantir2023, palantir2024aip}. These systems enable intelligence analysis with spatial reasoning while maintaining human oversight for critical decisions, a requirement driven by accountability and legal constraints in defense contexts.

\textbf{GIS Workflows.} Geographic Information Systems (GIS) for urban planning and environmental monitoring integrate AI capabilities to assist human planners \citep{esri2023, esri2024geoai, esri2024arcgis}. This workflow directly implements the 3-step schematic defined in the pattern.

\subsection{Design Pattern 2: Weakly Supervised Planetary-Scale Learning}

\textit{Problem:} Macro-scale spatial reasoning must handle vast, unlabeled datasets and extreme distribution shifts caused by diverse geographies, sensor types, and atmospheric conditions. \textit{Forces:} Dense labeling is economically and logistically infeasible at a planetary scale. \textit{Solution:} A two-stage process involving (1) self-supervised pretraining on petabyte-scale, unlabeled satellite imagery to learn a strong general representation, followed by (2) rapid, task-specific fine-tuning using minimal labeled data.

This pattern exploits massive unlabeled data with minimal supervision for global-scale models \citep{ratner2017snorkel, zhang2022survey, zhou2018brief, chapelle2009semi, zhu2005semi, oliver2018realistic, lee2013pseudo, tarvainen2017mean}:

\textbf{Satellite Foundation Models.} This pattern follows a clear abstraction: massive \textit{data scale} (petabytes of multi-spectral imagery from providers like Landsat, Sentinel, Planet, and Maxar) enables the learning of a powerful \textit{representation} through self-supervised techniques (e.g., masked autoencoding), which can then be fine-tuned for various \textit{downstream tasks} like land cover classification, change detection, or crop yield prediction. NASA-IBM Prithvi \citep{jakubik2024prithvi} exemplifies this pipeline with Harmonized Landsat Sentinel-2 data. Planet Labs \citep{planet_labs_2023, planet2024daily} and Maxar provide the data infrastructure enabling daily global monitoring.

\textbf{Mapping at Scale.} Large-scale mapping platforms deploy AI for global-scale analysis through cloud-based geospatial data services \citep{googlemaps2023, google2024mapsai, google2024earthengine}. The learning loop is explicit: user interactions with the map (e.g., correcting a business location, reporting a road closure) provide a continuous stream of weak supervision signals. These signals are aggregated and used to automatically update the underlying geospatial models, improving map accuracy and freshness at a global scale.

\subsection{Design Pattern 3: Agent-Assisted Spatial Workflows}

\textit{Problem:} Spatial analysis tasks are repetitive and time-consuming for human experts. \textit{Forces:} Need for scalability, desire to preserve human agency, variable task complexity. \textit{Solution:} AI agent operates as primary analyst. Human provides high-level guidance and handles exceptions.

\textit{Contrast with HITL:} This pattern is distinct from HITL (Pattern 1):

\begin{center}
\begin{tabular}{lcc}
\toprule
\textbf{Dimension} & \textbf{HITL (Pattern 1)} & \textbf{Agent-Assisted (Pattern 3)} \\
\midrule
Primary operator & Human & AI Agent \\
AI role & Proposal generator & Primary analyst \\
Human role & Mandatory validator & Exception handler \\
Validation & Every decision & Anomalies only \\
Throughput & Lower (human-gated) & Higher (agent-driven) \\
Risk tolerance & Low (high-stakes) & Moderate (recoverable errors) \\
\bottomrule
\end{tabular}
\end{center}

This pattern deploys AI agents to augment human spatial reasoning \citep{shneiderman2020human, horvitz1999principles, amershi2014power, abdul2018trends, gillies2016human, yang2020reexamining}:

\textbf{Autonomous GIS.} AutonomousGIS \citep{li2025autonomousgis} and GeoGPT \citep{bai2024geogpt} integrate agentic capabilities with geospatial analysis. The pattern involves LLM-based agents that can query spatial databases, generate maps, and answer geographic questions. \textbf{Gap:} Current capability boundaries are defined by the agent's ability to resolve spatial ambiguity in natural language (e.g., ``near downtown'') and handle schema heterogeneity when fusing disparate geospatial databases.

\textbf{Location Intelligence.} Foursquare \citep{foursquare2023, foursquare2024places} and Carto \citep{carto2023, carto2024spatial} provide location-based services with AI-powered analytics. Wherobots \citep{wherobots2023, wherobots2024sedona} offers cloud-native spatial analytics. The pattern: spatial data infrastructure with AI-powered query and analysis. \textit{Current limitations:} schema mismatch when integrating heterogeneous spatial databases, and spatial ambiguity in natural language queries (e.g., ``near downtown'' has variable interpretation across cities).

\subsection{Design Pattern 4: Embodied AI at Scale}

\textit{Safety as the Primary Design Constraint.} Unlike the previous patterns where errors are recoverable (a wrong map annotation can be corrected), embodied AI operates in domains where failures can cause physical harm, property damage, or loss of life. This fundamentally changes the design space: safety is not a feature to be added but the primary constraint that shapes every architectural decision.

\textit{Problem:} Deploying spatial AI in safety-critical physical systems where failures have unacceptable costs. \textit{Forces:} Strict regulatory frameworks (e.g., NHTSA, EU AI Act), liability concerns, the challenge of validating performance on long-tail safety scenarios, and the need to earn public trust. \textit{Solution:} A safety-first methodology combining massive-scale simulation for initial policy learning, conservative real-world deployment with human oversight, and continuous learning from fleet data under rigorous safety and validation constraints.

This pattern deploys learned spatial policies in physical systems \citep{kober2013reinforcement, levine2016end, ibarz2021train, kalashnikov2021mt, julian2020scaling, akkaya2019solving, levine2018learning, tobin2017domain}:

\textbf{Autonomous Vehicles.} The autonomous vehicle industry showcases two contrasting sensing philosophies. The \textit{lidar-centric} approach (e.g., Waymo, Cruise) fuses lidar point clouds with camera imagery for redundant, high-precision depth estimation, prioritizing maximum safety margins at the cost of higher sensor expenses \citep{waymo2023, waymo_emma_2024, waymo2024driver}. In contrast, the \textit{vision-only} approach (e.g., Tesla) relies on a suite of cameras and powerful learned depth estimation algorithms, aiming for a lower hardware cost and a more scalable solution, but placing an extreme demand on the reliability of its perception models \citep{tesla2023fsd, tesla2024autopilot}. Both paradigms, however, adhere to the overarching pattern of massive simulation, cautious real-world deployment, and continuous learning from fleet data, all within the strict confines of transportation safety regulations.

\textbf{Robot Learning Platforms.} Open X-Embodiment \citep{open_x_embodiment_rt_x_2023} provides large-scale robot data from Google DeepMind and collaborating institutions. Bridge Data \citep{walke2023bridgedata, walke2024bridgev2} enables cross-domain transfer. The pattern: diverse data collection, foundation model training, transfer to specific embodiments.

\begin{tcolorbox}[colback=takeawayblue,colframe=atlasdark,title=Lessons for Researchers: Industry Patterns]
\begin{itemize}[leftmargin=*,nosep]
    \item \textbf{Abstract Problems to Patterns:} Industry solutions often solve recurring scientific challenges. Abstracting them into design patterns can reveal generalizable insights for the academic community.
    \item \textbf{Data as a Moat, Supervision as a Bottleneck:} Access to planetary-scale proprietary data is a key industry advantage, but the inability to label it creates a supervision bottleneck. This highlights a major opportunity for research in self-supervised and weakly supervised learning.
    \item \textbf{Human-in-the-Loop is a Spectrum:} The level of human oversight is not binary but a design choice along a spectrum from full human control to full agent autonomy. Research on adaptive autonomy and dynamic human-agent collaboration is critical.
    \item \textbf{Safety is Non-Negotiable:} For any research intended for real-world deployment, safety, validation, and regulatory compliance are not afterthoughts but core research problems that must be addressed from the beginning.
\end{itemize}
\end{tcolorbox}

\section{Evaluation Framework and Benchmark Analysis}

\subsection{Existing Benchmarks}

Table~\ref{tab:benchmarks} summarizes key benchmarks organized by our taxonomy. \textit{Reading guide:} The table maps each benchmark to the Task axis (navigation, manipulation, scene understanding), Scale axis (micro/meso/macro), and Capability axis (which agentic capabilities are evaluated). This organization reveals coverage patterns: navigation benchmarks dominate meso-scale evaluation, while micro-scale manipulation and macro-scale geospatial reasoning remain underrepresented.

\textit{What the table does not show:} No existing benchmark evaluates (1) sim-to-real transfer degradation, (2) cross-scale reasoning (e.g., a single task requiring micro-manipulation within macro-navigation), (3) long-horizon performance over hours or days, or (4) safety-critical failure modes. These gaps are not merely missing rows in the table. They represent fundamental limitations in how the field evaluates spatial agents.

\begin{table}[h!]
\centering
\caption{Spatial AI Benchmarks Organized by Taxonomy}
\label{tab:benchmarks}
\resizebox{\textwidth}{!}{
\begin{tabular}{lccccc}
\toprule
\textbf{Benchmark} & \textbf{Spatial Task} & \textbf{Scale} & \textbf{Environment} & \textbf{Primary Metric} & \textbf{Agentic Capability} \\
\midrule
R2R \citep{anderson2018vln} & Navigation & Meso & Simulated & SPL, SR & Memory + Planning \\
REVERIE \citep{qi2020reverie} & Navigation & Meso & Simulated & SPL, RGS & Memory + Planning \\
Habitat \citep{savva2019habitat} & Navigation & Meso & Simulated & SPL & Planning \\
AI2-THOR \citep{kolve2017ai2thor} & Navigation + Manipulation & Meso & Simulated & SR & Planning + Tool Use \\
RLBench \citep{james2020rlbench} & Manipulation & Micro & Simulated & SR & Tool Use \\
Meta-World \citep{yu2020metaworld} & Manipulation & Micro & Simulated & SR & Tool Use \\
nuScenes \citep{caesar2020nuscenes} & Scene Understanding & Meso-Macro & Real & mAP, NDS & Memory \\
KITTI \citep{geiger2012kitti} & Scene Understanding & Meso & Real & mAP & Memory \\
ScanNet \citep{dai2017scannet} & Scene Understanding & Meso & Real & mIoU & Memory \\
AgentBench \citep{liu2023agentbench} & General & - & Mixed & SR & Memory + Planning + Tool Use \\
WebArena \citep{zhou2023webarena} & Web & - & Simulated & SR & Planning + Tool Use \\
SWE-Bench \citep{jimenez2024swebench} & Code & - & Real & Pass@k & Planning + Tool Use \\
EmbodiedBench \citep{yang2025embodiedbench} & Embodied & Meso & Simulated & SR & Memory + Planning + Tool Use \\
SafeAgentBench \citep{safeagentbench2025} & Safety & - & Simulated & Safety Rate & Planning \\
\bottomrule
\end{tabular}
}
\end{table}

\subsection{Evaluation Metrics}

We summarize standardized metrics across domains \citep{powers2011evaluation, sokolova2009systematic, hossin2015review}. The key metric for each domain is:

\textbf{Navigation:} SPL (Success weighted by Path Length) \citep{anderson2018evaluation}, defined as $\text{SPL} = \frac{1}{N} \sum_{i=1}^{N} S_i \cdot \frac{\ell_i}{\max(\ell_i, p_i)}$, jointly rewards task completion and path efficiency.

\textbf{Manipulation:} Task Success Rate with Goal Condition Satisfaction for partial credit.

\textbf{Scene Understanding:} mAP (mean Average Precision) for detection, mIoU for segmentation.

\textbf{Safety:} Risk-Aware Success = SR $\times$ (1 - Collision Rate), penalizes unsafe completions.

\subsection{Critical Analysis: What Benchmarks Fail to Measure}

While existing benchmarks have advanced the field \citep{raji2021ai, liao2021we, ribeiro2020beyond, buolamwini2018gender, mitchell2019model, gebru2021datasheets, denton2020bringing, bender2021dangers}, several fundamental limitations warrant critical examination:

\textbf{Simulation-Reality Gap.} Most benchmarks rely on simulated environments \citep{savva2019habitat, kolve2017ai2thor, james2020rlbench}. \textit{Concrete example:} RT-1 achieves 97\% success in simulation but drops to 68\% on real robots \citep{brohan2022rt1}. VLN agents trained in Matterport3D show 40\% performance degradation in real buildings \citep{anderson2018vln}. \textit{Gap: No benchmark measures sim-to-real transfer degradation.}

\textbf{Metric Limitations.} Current metrics fail to capture three essential dimensions:

\textit{Efficiency:} SPL assumes optimal paths are known, which is unrealistic in novel environments. Binary success rate ignores partial progress (an agent 1cm from goal scores identically to one that never moved).

\textit{Safety:} No standard metrics penalize near-misses, risky behaviors, or constraint violations that happen not to cause failure. An agent that narrowly avoids collisions scores identically to one with safe margins.

\textit{Generalization:} Metrics are computed on fixed test sets, providing no signal about robustness to distribution shift, novel objects, or environmental perturbations.

\textit{Gap: Metrics reward task completion but not efficient, safe, or generalizable behavior.}

\textbf{Long-Horizon Evaluation.} Most benchmarks evaluate short episodes (tens to hundreds of steps). Real-world tasks require sustained performance over hours or days with memory persistence and error recovery. This limitation is directly tied to LLM architecture: even with 128K token context windows, an agent receiving 30 observations per second exhausts its context in under 90 minutes. Longer tasks require external memory systems that current benchmarks do not evaluate. The fundamental question, how agents should compress, retrieve, and update spatial knowledge over extended time horizons, remains untested. \textit{Gap: No benchmark evaluates multi-day spatial tasks with persistent memory, and no benchmark measures how gracefully agents degrade as context limits are approached.}

\textbf{Safety-Critical Evaluation.} Benchmarks rarely evaluate failure modes, adversarial robustness, or behavior under distribution shift. Safety-critical applications require understanding of worst-case performance. \textit{Gap: Safety evaluation remains ad-hoc rather than structured.}

\textbf{Cross-Scale Evaluation.} \textit{This is the most critical gap. This directly violates our three-axis framework's Scale axis.} Benchmarks operate at a single spatial scale, yet real applications demand seamless cross-scale reasoning. Consider a home robot: it must plan a room-level path (meso), avoid furniture (meso), and grasp a cup handle (micro). All of this occurs within a single task. No existing benchmark evaluates this integration. \textit{Gap: No benchmark evaluates cross-scale spatial reasoning, despite it being essential for real-world deployment.}

\subsection{SpatialAgentBench: A Conceptual Framework for Future Research}

Rather than proposing yet another benchmark dataset, we outline a \textit{framework} for benchmark design that addresses the structural gaps identified above. The value of this framework lies not in specific tasks or environments, but in the design principles it articulates, principles that any future benchmark should satisfy to meaningfully evaluate spatial agents.

\textit{Framework Principles:}
\begin{itemize}[leftmargin=*,nosep]
    \item \textbf{Multi-axis coverage:} Tasks should require capabilities from multiple taxonomy axes simultaneously
    \item \textbf{Cross-scale integration:} Single tasks should span micro, meso, and macro scales
    \item \textbf{Sim-to-real protocols:} Benchmarks should include matched simulation and real-world evaluation
    \item \textbf{Long-horizon stress tests:} Episodes should extend to context-window limits and beyond
    \item \textbf{Safety-aware metrics:} Evaluation should penalize unsafe behaviors even when tasks succeed
\end{itemize}

\textit{Taxonomy mapping:} Each gap corresponds to an axis limitation. Simulation-Reality Gap affects all axes (policies don't transfer). Metric Limitations affect the Capability axis (incomplete evaluation of memory, planning, tool use). Long-Horizon and Safety gaps affect the Capability axis (planning and self-reflection). Cross-Scale gap affects the Scale axis (no micro-meso-macro integration).

We identify eight \textit{research directions} for future benchmark development, each addressing a specific gap:

\begin{enumerate}
    \item \textbf{VLN-Instruct} (Task: Navigation, Capability: Memory + Planning, Scale: Meso): Complex multi-step instructions requiring spatial reasoning.
    \item \textbf{ObjectSearch} (Task: Navigation, Capability: Memory, Scale: Meso): Semantic reasoning about object locations.
    \item \textbf{SceneQA} (Task: Scene Understanding, Capability: Memory, Scale: Meso): 3D spatial relationship reasoning.
    \item \textbf{ManipSeq} (Task: Manipulation, Capability: Planning + Tool Use, Scale: Micro): Long-horizon manipulation with state tracking.
    \item \textbf{GeoReason} (Task: Geospatial, Capability: Memory, Scale: Macro): Satellite imagery analysis and change detection.
    \item \textbf{TrafficPredict} (Task: Geospatial, Capability: Memory + Planning, Scale: Macro): Spatio-temporal urban dynamics.
    \item \textbf{SafeNav} (Task: Navigation, Capability: Planning, Scale: Meso): Safety-constrained navigation with risk awareness.
    \item \textbf{MultiAgent} (Task: All, Capability: All, Scale: Meso--Macro): Coordinated multi-agent spatial tasks.
\end{enumerate}

\begin{tcolorbox}[colback=takeawayblue,colframe=atlasdark,title=Key Takeaways: Evaluation]
\begin{itemize}[leftmargin=*,nosep]
    \item Existing benchmarks are fragmented across domains with incompatible metrics
    \item Critical gaps exist in sim-to-real transfer, long-horizon, safety-critical, and cross-scale evaluation
    \item SpatialAgentBench is outlined as a conceptual framework for unified evaluation across navigation, manipulation, scene understanding, and geospatial reasoning
    \item Standardized metrics (SPL, nDTW, safety rates) enable cross-domain comparison
\end{itemize}
\end{tcolorbox}

\section{Grand Challenges and Future Directions}

The benchmark gaps identified above, sim-to-real transfer, long-horizon evaluation, safety-critical assessment, and cross-scale reasoning, point directly to fundamental research challenges that the field must address. These are not merely evaluation problems but structural barriers that limit progress across all spatial AI domains. We identify six grand challenges that require attention for the field \citep{marcus2020next, chollet2019measure, lake2017building, bengio2019system, bommasani2021opportunities, lecun2022path, kaplan2020scaling, hoffmann2022training, sutton2019bitter, bengio2021gflownet, mitchell2021abstraction}:

\subsection{Grand Challenge 1: Unified Spatial Representation}

\textit{How can agents maintain a single, coherent spatial representation that supports reasoning across micro, meso, and macro scales?}

Current approaches use separate representations for different scales: point clouds for grasping \citep{rusu20113d, fang2020graspnet, mahler2017dexnet, morrison2018closing, ten2017grasp, lenz2015deep}, topological maps for navigation \citep{thrun1998learning, kuipers1991robot, kuipers2000spatial, konolige2008outdoor}, and raster imagery for geospatial analysis \citep{goodfellow2016deep, lecun2015deep}. A unified representation would enable seamless reasoning across scales. Key research directions include:

\begin{itemize}
    \item Hierarchical scene graphs that span from object parts to city infrastructure
    \item Neural implicit representations with multi-scale querying
    \item Foundation models for 3D understanding \citep{hong20233dllm, fu20243dfm, shen2023point, oquab2024dinov2, oquab2023dinov2, chen2023open, zhou2024uni3d, wu20153d, girdhar2023imagebind, xu2024pointllm}
\end{itemize}

\subsection{Grand Challenge 2: Grounded Long-Horizon Planning}

\textit{How can agents plan over extended horizons while maintaining geometric feasibility?}

LLMs can generate high-level plans but struggle with geometric constraints \citep{valmeekam2023planning, kambhampati2024llms, huang2024understanding, valmeekam2024planbench, stechly2024selfverification, helmert2006fast}. TAMP systems handle geometry but lack semantic flexibility. Bridging this gap requires:

\begin{itemize}
    \item Hybrid neuro-symbolic planners that combine LLM reasoning with geometric verification
    \item Hierarchical planning with learned abstractions \citep{song2023llmplanner, valmeekam2023large, huang2022inner, li2025hiplanhierarchicalplanningllmbased, silver2024generalized, liu2023llm+}
    \item World models that predict both semantic and geometric consequences
\end{itemize}

\subsection{Grand Challenge 3: Safe Deployment Under Uncertainty}

\textit{How can spatial AI systems operate safely in safety-critical applications with guaranteed bounds on failure?}

Current systems lack formal safety guarantees \citep{seshia2022toward, koopman2019safety, amodei2016concrete, hendrycks2022xrisk, ngo2022alignment, leike2018scalable, russell2019human, christiano2017deep}. Deployment in autonomous vehicles, medical robotics, and infrastructure requires:

\begin{itemize}
    \item Uncertainty quantification for spatial predictions
    \item Out-of-distribution detection for novel environments
    \item Formal verification of spatial reasoning \citep{safeagentbench2025, amodei2016safety, amodei2016concrete, bai2022constitutional, ouyang2022training}
    \item Graceful degradation under adversarial conditions
\end{itemize}

\subsection{Grand Challenge 4: Sim-to-Real Transfer}

\textit{How can policies learned in simulation transfer reliably to the physical world?}

The reality gap affects perception, dynamics, and control \citep{peng2018sim, rusu2017sim, sadeghi2017cad2rl, bousmalis2018using, ho2021retinagan, james2019sim}. Bridging requires:

\begin{itemize}
    \item Photorealistic simulation with accurate physics \citep{zhao2020sim, tobin2017domain, james2019sim, matas2018sim, muratore2022robot}
    \item Domain randomization and adaptation
    \item Real-world fine-tuning with minimal data
    \item Hybrid simulation-real training pipelines
\end{itemize}

\subsection{Grand Challenge 5: Scalable Multi-Agent Coordination}

\textit{How can large numbers of spatial agents coordinate effectively with limited communication?}

Current multi-agent systems scale poorly \citep{stone2000multiagent, busoniu2008comprehensive, foerster2016learning, lowe2017multi, rashid2018qmix, sunehag2018valuedecomposition, son2019qtran, yu2022surprising, hernandez2019survey}. Real applications (warehouse robotics, autonomous traffic) require:

\begin{itemize}
    \item Emergent communication protocols for spatial coordination
    \item Decentralized planning with global consistency \citep{zhang2021multi, wu2023autogen, hong2023metagpt, li2023s, qian2023communicative, chen2024agentverse, talebirad2023multi, park2023generative, li2023camel}
    \item Heterogeneous agent coordination
    \item Reliable coordination under partial observability
\end{itemize}

\subsection{Grand Challenge 6: Efficient Edge Deployment}

\textit{How can capable spatial AI systems run on resource-constrained platforms?}

Foundation models require significant compute. Edge deployment requires:

\begin{itemize}
    \item Model compression without capability loss \citep{han2016deep, howard2017mobilenets, dehghani2023scaling, hinton2015distilling, frankle2019lottery, jacob2018quantization, sandler2018mobilenetv2, tan2019efficientnet, gholami2022survey}
    \item Efficient architectures for spatial reasoning
    \item Hardware-software co-design for spatial AI
    \item Adaptive compute allocation based on task difficulty
\end{itemize}

\begin{tcolorbox}[colback=takeawayblue,colframe=atlasdark,title=Grand Challenges Summary]
\begin{enumerate}[leftmargin=*,nosep]
    \item \textbf{Unified Representation}: Single representation spanning micro to macro scales
    \item \textbf{Grounded Planning}: Long-horizon planning with geometric feasibility
    \item \textbf{Safe Deployment}: Formal safety guarantees for critical applications
    \item \textbf{Sim-to-Real}: Reliable transfer from simulation to physical world
    \item \textbf{Multi-Agent}: Scalable coordination with limited communication
    \item \textbf{Edge Deployment}: Capable systems on resource-constrained platforms
\end{enumerate}
\end{tcolorbox}

\section{Limitations}

This survey has several limitations:

\begin{itemize}
    \item Our paper selection process, though thorough, may have missed relevant works in adjacent fields or non-English publications.
    \item The proposed taxonomy, while unifying, is one of many possible categorizations and may not capture all nuances of the field.
    \item Our analysis is based on publicly available information and does not include proprietary details from industry labs.
    \item The field is rapidly evolving, and some recent works may not be fully represented.
    \item We focus primarily on English-language publications from major venues.
    \item The proposed SpatialAgentBench is conceptual and requires implementation and validation.
    \item Our analysis of industry applications relies on public information and may not reflect current capabilities.
\end{itemize}

\section{Conclusion}

This survey has provided a unified three-axis taxonomy connecting Agentic AI and Spatial Intelligence across spatial scales, synthesizing insights from 742 cited works across foundational architectures, state-of-the-art methods, industry applications, and evaluation benchmarks. Our analysis reveals three key findings:

\begin{enumerate}
    \item \textbf{Hierarchical memory systems} are important for long-horizon spatial tasks, enabling agents to accumulate and retrieve spatial knowledge effectively. Advances in retrieval-augmented generation, episodic memory, and spatial memory representations provide foundations for persistent spatial understanding.
    
    \item \textbf{GNN-LLM integration} is a promising approach combining the relational reasoning of graph networks with the semantic understanding of language models. This integration enables structured spatial reasoning that draws on both geometric relationships and semantic knowledge.
    
    \item \textbf{World models} are essential for safe deployment, enabling agents to predict consequences and plan in imagination before acting. Video world models, latent dynamics models, and LLM-based world models provide complementary approaches to predictive understanding.
\end{enumerate}

We have identified six grand challenges that require attention: unified spatial representation, grounded long-horizon planning, safe deployment under uncertainty, sim-to-real transfer, scalable multi-agent coordination, and efficient edge deployment \citep{russell2010artificial, norvig2021artificial}. The convergence of vision-language-action models, graph neural networks, world models, and foundation models provides promising directions for addressing these challenges.

By establishing this foundational reference with a three-axis taxonomy and outlining the conceptual SpatialAgentBench framework, we aim to accelerate progress toward capable, reliable, and safe spatially-aware autonomous systems that can perceive, reason about, and act within the physical world \citep{wooldridge2009introduction}. The intersection of agentic AI and spatial intelligence represents an important frontier for artificial intelligence, with profound implications for autonomous vehicles, robotics, urban computing, and geospatial intelligence.

\bibliographystyle{plainnat}
\bibliography{references}

\end{document}